%% file: main.tex
\documentclass[10pt,twocolumn,letterpaper]{article}

\usepackage{cvpr}              

\input{preamble}

\definecolor{cvprblue}{rgb}{0.21,0.49,0.74}
\usepackage[pagebackref,breaklinks,colorlinks,allcolors=magenta]{hyperref}
\usepackage{amssymb,bbding,pifont}

\definecolor{myblue}{RGB}{66,133,244}
\definecolor{mygreen}{RGB}{51,168,83}
\definecolor{myyellow}{RGB}{251,188,3}
\definecolor{myred}{RGB}{234,67,53}
\definecolor{mygrey}{RGB}{95,99,104}
\definecolor{mypup}{RGB}{153,0,204}

\def\paperID{*****} 
\def\confName{CVPR}
\def\confYear{2025}
\hypersetup{breaklinks,colorlinks}
\title{LoViF 2026 Challenge on Human-oriented Semantic Image Quality Assessment: Methods and Results}

\author{
Xin Li$^{\dagger}$\quad Daoli Xu$^{\dagger}$\quad Wei Luo$^{*}$\quad Guoqiang Xiang$^{\dagger}$\quad Haoran Li$^{*}$\quad Chengyu Zhuang$^{*}$\quad Zhibo Chen$^{\dagger}$\\ Jian Guan$^{\dagger}$\quad Weiping Li$^{\dagger}$ \\
Weixia Zhang\quad Wei Sun\quad Zhihua Wang\quad Dandan Zhu\quad Chengguang Zhu\quad Ayush Gupta\\ Rachit Agarwal\quad Shouvik Das\quad Biplab Ch Das\quad Amartya Ghosh\quad Kanglong Fan\quad Wen Wen\\ Shuyan Zhai\quad Tianwu Zhi\quad Aoxiang Zhang\quad Jianzhao Liu\quad Yabin Zhang\\ Jiajun Wang\quad Yipeng Sun\quad Kaiwei Lian\quad Banghao Yin}

\begin{document}
\maketitle
\renewcommand{\thefootnote}{}
\footnotetext{$^{\dagger}$X. Li (xin.li@ustc.edu.cn), D. Xu,  G. Xiang, Z. Chen, J. Guan, W. Li are the challenge organizers. The contact details can be found in section~\ref{organizers}. $^{*}$W. Luo, H. Li, and C. Zhuang are support staff of this competition. 
}
\footnotetext{The other authors are participants of the LoViF 2026 Challenge on Human-oriented Semantic Image Quality Assessment.}
\footnotetext{The Competition website and the SeIQA database~\url{https://www.codabench.org/competitions/13692/}}
\footnotetext{The LoViF 2026 website:~\url{lovif-cvpr2026-workshop.github.io}}

\input{sec/0_abstract}    
\input{sec/1_intro}

\input{sec/2_Challenge}
\input{sec/3_Challenge_Results}
\input{sec/4_Teams_and_Methods_supp}
\input{sec/5_Appendix}

\section*{Acknowledgments}
This challenge is supported by EMERGETECH and  the Joint Laboratory of EMERGETECH and Institute of Advanced Technology, University of Science and Technology of China, on Intelligent Media Computing.  This work is also partially supported by the Postdoctoral Fellowship Program of CPSF under Grant Number GZC20252293 and the Fundamental Research Funds for the Central Universities (No. WK2100250064).

{
    \small
    \bibliographystyle{ieeenat_fullname}
    \bibliography{main}
}


\end{document}

%% file: preamble.tex
\usepackage[dvipsnames]{xcolor}


%% file: sec/0_abstract.tex
\begin{abstract}
This paper reviews the LoViF 2026 Challenge on Human-oriented Semantic Image Quality Assessment. This challenge aims to raise a new direction, \textit{i.e.,} how to evaluate the loss of semantic information from the human perspective, intending to promote the development of some new directions, like semantic coding, processing, and semantic-oriented optimization, etc. Unlike existing datasets of quality assessment, we form a dataset of human-oriented semantic quality assessment, termed the SeIQA dataset. This dataset is divided into three parts for this competition: (i) training data: 510 pairs of degraded images and their corresponding ground truth references;
(ii) validation data: 80 pairs of degraded images and their corresponding ground-truth references; (iii) testing data: 160 pairs of degraded images and their corresponding ground-truth references. The primary objective of this challenge is to establish a new and powerful benchmark for human-oriented semantic image quality assessment. There are a total of 58 teams registered in this competition, and 6 teams submitted valid solutions and fact sheets for the final testing phase. These submissions achieved state-of-the-art (SOTA) performance on the SeIQA dataset. 
\end{abstract}

%% file: sec/1_intro.tex
\section{Introduction}
\label{sec:intro}
Image quality assessment has traditionally focused on perceptual fidelity~\cite{MANIQA,CLIPIQA,DBCNNQA,guan2024qmamba,metaiqaQA,li2023freqalign,liu2020lira, zhai2020perceptualQA,ji2019blindsemantics}, \emph{i.e.}, whether an image is sharp, natural, and visually pleasing to human observers. While such a perspective is essential for conventional restoration, compression, and enhancement tasks, it is no longer sufficient for many emerging scenarios in the era of generative models and intelligent visual systems~\cite{AIGCIQA2023QA,AGIQA3KQA,feng2025diffsemantics,yang2024videosemantics}. On the other hand, existing semantic quality evaluation is still largely indirect, often relying on downstream recognition or task-specific performance as a proxy~\cite{yang2024videosemantics,lu2024preprocessing,SR1}. Such evaluation protocols are typically limited in generality and may not align well with human semantic understanding.

This motivates the study of human-oriented semantic quality assessment, whose goal is to measure image quality from the human perspective of semantic preservation rather than human perceptual fidelity or machine semantic quality. Semantic quality is more closely related to whether key objects, attributes, relations, and scene-level meanings are preserved after visual processing. This problem is highly relevant to a broad range of applications, including semantic-aware visual coding, transmission, enhancement, and AI-generated content analysis~\cite {li2021task,huang2024vbench}. For instance, in many practical scenarios, users may care more about preserving the semantic information they focus on than about maintaining all low-level details. However, the existing semantic quality evaluation for machines is narrowly related to the tasks performed by machines. Therefore, it becomes increasingly important to evaluate whether the conveyed visual content remains semantically faithful to what human regard as the essential meaning of the scene.

To bridge this gap, we organized the LoViF 2026 First Challenge on Human-oriented Semantic Quality Assessment and introduced a new benchmark for this problem, which we refer to as the \textit{\textbf{SeIQA}} dataset in this report. The challenge is intended to facilitate a reliable human-oriented semantic quality assessment benchmark that is more deeply aligned with human semantic understanding, thereby promoting future research on semantics-oriented visual coding, transmission, and processing. Different from conventional IQA benchmarks~\cite{AIGCIQA2023QA,lu2024kvqQA,li2024ntireQA} that mainly emphasize perceptual distortions such as blur, noise, or compression artifacts, the proposed benchmark explicitly targets semantic consistency between degraded images and their references, especially in challenging cases where perceptual appearance and semantic correctness may not be well aligned.

From the benchmark perspective, the released training set contains 510 pairs of degraded images and their corresponding reference images. Participants are required to predict semantic quality scores, and the final ranking is determined by a weighted combination of Spearman rank-order correlation coefficient (SROCC) and Pearson linear correlation coefficient (PLCC). This protocol encourages methods that are not only well correlated with human semantic judgments in ranking, but also accurate in absolute score prediction. We expect this challenge to serve as a useful starting point for building a more human-centered semantic quality assessment framework and to stimulate future advances in semantics-aware visual processing.

This challenge is held with the LoViF Workshop~\footnote{\url{https://lovif-cvpr2026-workshop.github.io/}}, containing this challenge and a series of other challenges on: real-world all-in-one image restoration~\cite{lovif2026realir}, efficient VLM for multimodal creative quality scoring~\cite{lovif2026MQualityScoring}, weather removal in videos~\cite{lovif2026WeatherRemoval} and holistic quality assessment for 4D world model~\cite{lovif2026HQA}.

%% file: sec/2_Challenge.tex
\section{Challenge}
\label{sec:challenge}

The LoViF 2026 First Challenge on Semantic Quality Assessment aims to advance the study of human-oriented semantic quality assessment beyond conventional perceptual fidelity. While existing image quality assessment benchmarks mainly emphasize whether an image is visually pleasing or perceptually faithful, this challenge focuses on a different and increasingly important question: whether the degraded/decoded image still preserves the semantic content that humans consider essential for understanding the scene. To support this purpose, the challenge introduces a dedicated benchmark for human-oriented semantic quality assessment, which we refer to as the \textit{SeIQA} benchmark. Different from traditional IQA settings that mainly measure low-level distortions such as blur, noise, and compression artifacts, the proposed benchmark emphasizes whether image degradation changes the semantics conveyed by visual content. This setting is particularly relevant to semantics-oriented visual coding, transmission, and processing, where preserving semantically important content may be more critical than maintaining all low-level details.

The released benchmark contains paired degraded and reference images for model development, together with held-out validation and testing data for official evaluation. Specifically, the training set contains 510 pairs of degraded images and their corresponding reference images. This training set is constructed and labeled by professional workers and an intelligent application, as DouBao\footnote{\url{https://www.doubao.com}}. We also construct the validation dataset and testing dataset, resulting in 80 and 160 pairs of degraded images and reference images. These datasets are accurately labeled and averaged by 30 humans. By constructing the challenge in this way, the benchmark provides a first dedicated testbed for studying semantic quality prediction under human-oriented criteria.

The challenge is hosted on the Codabench platform~\cite{xu2022codabench} as part of the LoViF 2026 workshop at CVPR 2026. It is jointly organized by USTC and EMERGETECH, with the aim of establishing a forward-looking benchmark for human-oriented semantic quality assessment and promoting future research in this emerging area.
The challenge consists of two phases. (i) In the development phase, the validation set is released for online evaluation, and 21 teams submitted their results during this stage. (ii) In the test phase, the test set was released for final evaluation, and 22 teams submitted their final results on the Codabench platform. Among them, six teams further submitted valid fact sheets and were finally included in the official ranking.

\begin{table*}[tp]
    \centering
    \caption{Quantitative results of the LoViF 2026 The First Challenge on Human-oriented Semantic Quality Assessment.  The best and second-best values per column are highlighted in \textcolor{red}{red} and \textcolor{blue}{blue}, respectively.  "Params." and "GFlops" are reported from the teams’ factsheets.  A checkmark ($\checkmark$) indicates the use of ensembles or extra data, while a cross ($\otimes$) indicates otherwise.}
    \resizebox{\textwidth}{!}{
    \begin{tabular}{cc|cccc|cc|c}
    \toprule
    Team & Leader & Final Score $\uparrow$ & PLCC$\uparrow$ & SROCC$\uparrow$ & Runtime per images(s) $\downarrow$ & GPU& Extra Data & Rank\\ \midrule
    RedpanQA Alliance & Weixia~Zhang & \textcolor{red}{0.8724} & \textcolor{red}{0.8764} & \textcolor{red}{0.8697} & 12.0 & $\checkmark$ & $\checkmark$ & 1\\
    Ayush Gupta & Ayush~Gupta & \textcolor{blue}{0.8711} & \textcolor{blue}{0.8763} & 0.8677& 5.0 & $\checkmark$ & $\checkmark$ & 2\\
    RuntimeTerror  &Rachit Agarwal  & 0.8693& 	0.8710& \textcolor{blue}{0.8681}& 1.0 & $\checkmark$ & $\otimes$ & 3\\
    QA-FTE & Kanglong~Fan & 0.8560& 0.8584& 0.8544& 12.0 & $\checkmark$ & $\checkmark$ & 4\\
    DSS-SQA & Jiajun~Wang & 0.8469 & 0.8418& 	0.8503& \textcolor{red}{	0.22} & $\checkmark$ & $\checkmark$ & 5\\
    cythdg  & Banghao Yin  & 0.8243& 0.8324& 0.8189& 	\textcolor{blue}{0.57} & $\checkmark$ & $\otimes$ & 6\\
    \bottomrule
    \end{tabular}}
    \label{tab:results}
\end{table*}

%% file: sec/3_Challenge_Results.tex
\section{Challenge Results}
The quantitative comparison is summarized in Table~\ref{tab:results}. Overall, team RedpanQA Alliance achieved the best performance in the challenge, obtaining the highest Final Score of 0.8724, together with the best PLCC of 0.8764 and the best SROCC of 0.8697. Team Ayush Gupta ranked second with a very competitive Final Score of 0.8711, only slightly lower than the winning team, and achieved a nearly identical PLCC of 0.8763, demonstrating strong prediction accuracy and robustness. Team RuntimeTerror placed third with a Final Score of 0.8693 and obtained the second-best SROCC of 0.8681, indicating strong rank consistency with human judgments.

It is worth noting that the top three teams achieved very close overall performance, with less than 0.004 difference in Final Score, which reflects the highly competitive nature of this challenge. Among them, team RedpanQA Alliance and Ayush Gupta both benefited from additional resources, while RuntimeTerror achieved comparable performance without using extra data, suggesting a favorable balance between effectiveness and methodological efficiency. Team QA-FTE ranked fourth with solid performance across both PLCC and SROCC, while team DSS-SQA and cythdg completed the leaderboard with Final Scores of 0.8469 and 0.8243, respectively.

In terms of efficiency, DSS-SQA achieved the fastest inference speed with a runtime of only 0.22 seconds per image, followed by cythdg with 0.57 seconds and RuntimeTerror with 1.0 second. In contrast, the top-ranked teams RedpanQA Alliance and QA-FTE required 12.0 seconds per image, while team Ayush Gupta required 5.0 seconds. These results reveal a trade-off between prediction accuracy and computational efficiency. In particular, team RuntimeTerror stands out as a strong compromise, achieving top-three performance with relatively low runtime and without relying on extra data. Overall, the results demonstrate that recent semantic quality assessment methods have made encouraging progress, while also highlighting that competitive performance can be achieved through different design choices in terms of accuracy, efficiency, and resource usage.

%% file: sec/4_Teams_and_Methods_supp.tex
\section{Teams and Methods}
\label{sec:teams_and_methods}

\subsection{Redpan QA Alliance}

\begin{figure*}[t]
    \centering
    \includegraphics[width=0.8\textwidth]{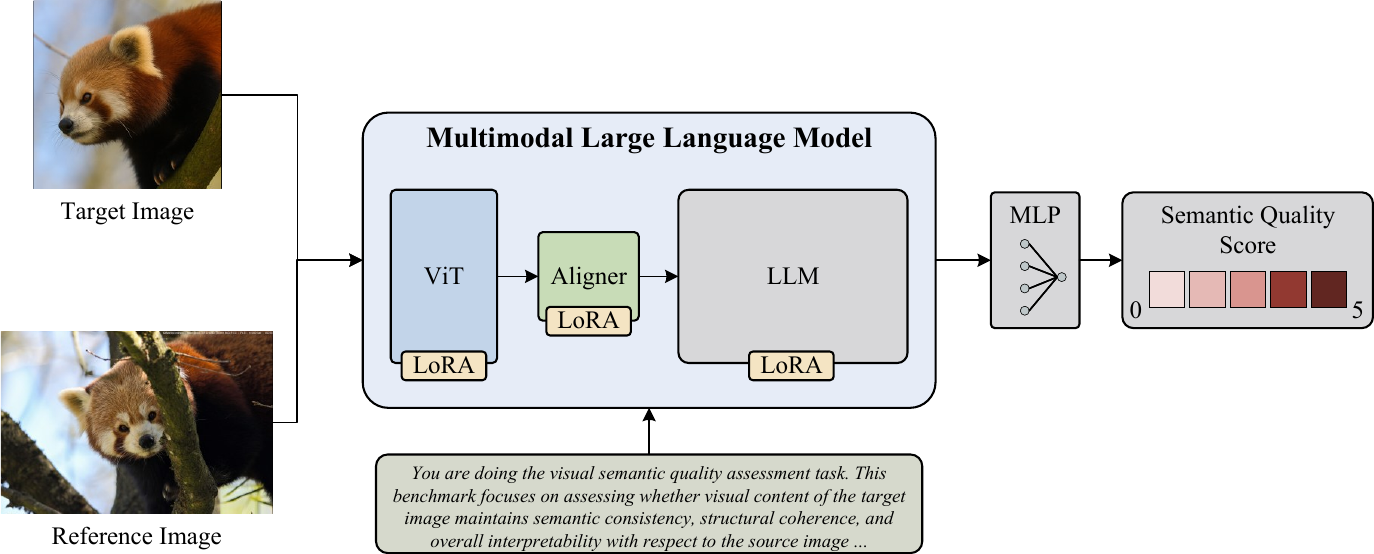}
    \caption{System diagram of Team Redpan QA Alliance}
    \label{fig:Redpan QA Alliance}
\end{figure*}

This team proposes a reference-based framework for semantic quality assessment based on multimodal large language models (MLLMs). As shown in Fig.~\ref{fig:Redpan QA Alliance}, given a source image, a target image, and a task-specific prompt, the model learns a joint multimodal representation that captures the semantic relationship between the two images under the defined assessment objective. Instead of generating textual outputs, the framework directly uses the hidden representations of the MLLM as high-level semantic features and employs a lightweight MLP regressor to predict a continuous quality score. For efficient adaptation, LoRA-based fine-tuning~\cite{hu2022lora} is adopted while keeping most pretrained parameters frozen.

Inspired by~\cite{zhang2021uncertainty, zhang2023continual, zhang2024task}, we train the model with a combination of a PLCC-induced loss and a fidelity loss~\cite{tsai2007frank}:
\begin{equation}
\mathcal{L}
=
\lambda_{1}\mathcal{L}_{\mathrm{Fid}}
+
\lambda_{2}\mathcal{L}_{\mathrm{PLCC}}.
\end{equation}

Given a mini-batch of predictions \(\hat y_i\) and ground-truth scores \(y_i\), where \(i=1,\dots,B\) indexes samples, the PLCC-induced loss is defined as
\begin{equation}
\mathcal{L}_{\mathrm{PLCC}}
=
1-
\frac{
\sum_{i=1}^{B}(\hat y_i-\bar{\hat y})(y_i-\bar y)
}{
\sqrt{\sum_{i=1}^{B}(\hat y_i-\bar{\hat y})^2}
\sqrt{\sum_{i=1}^{B}(y_i-\bar y)^2}
},
\end{equation}
where
\begin{equation}
\bar{\hat y}=\frac{1}{B}\sum_{i=1}^{B}\hat y_i,
\qquad
\bar y=\frac{1}{B}\sum_{i=1}^{B}y_i.
\end{equation}
This term encourages high linear correlation between predictions and subjective scores.

The fidelity loss is formulated on pairwise score differences:
\begin{equation}
\mathcal{L}_{\mathrm{Fid}}
=
\frac{1}{|\mathcal P|}
\sum_{(i,j)\in\mathcal P}
\Bigl[
1-
\bigl(
\sqrt{p_{ij} g_{ij}}
+
\sqrt{(1-p_{ij})(1-g_{ij})}
\bigr)
\Bigr],
\end{equation}
where
\begin{equation}
p_{ij}=\Phi(\hat y_i-\hat y_j),
\qquad
g_{ij}=\frac{\operatorname{sign}(y_i-y_j)+1}{2},
\end{equation}
and
\begin{equation}
\mathcal P=\{(i,j)\mid i<j\}.
\end{equation}
Here, \(\Phi(\cdot)\) denotes the standard Gaussian cumulative distribution function. This term enforces consistency between the predicted pairwise ordering and the ground-truth ordering.

\noindent\textbf{Training Details.} 
For training, the team fine-tuned the model for 1 or 3 epochs using LoRA with rank 64 and scaling factor \(\alpha=128\). We perform LoRA finetuning on all model components, \ie, visual encoder, visual aligner, and the large language model. The initial learning rate is set to \(1\times10^{-4}\) and updated with a cosine decay schedule. Training is conducted on two NVIDIA RTX A5880 Ada GPUs, with a per-GPU batch size of 16. During inference, min-max normalization is applied to calibrate the predicted scores to the target range of 0 to 5.

\noindent\textbf{Ensembles and fusion strategies.} 
For the final system, the team further employs an ensemble of three Qwen-based MLLM variants, specifically, Qwen3-VL-4B-Instruct~\cite{bai2025qwen3} (two sets of model weights trained for 1 and 3 epochs, respectively), Qwen3-VL-8B-Instruct~\cite{bai2025qwen3} (one model weight trained for 3 epochs). Each model follows the same representation learning and regression pipeline, and their outputs are aggregated to obtain the final prediction, which improves robustness and overall performance. 

\subsection{Ayush Gupta}

\begin{figure}[t]
    \centering
    \includegraphics[width=1.0\linewidth]{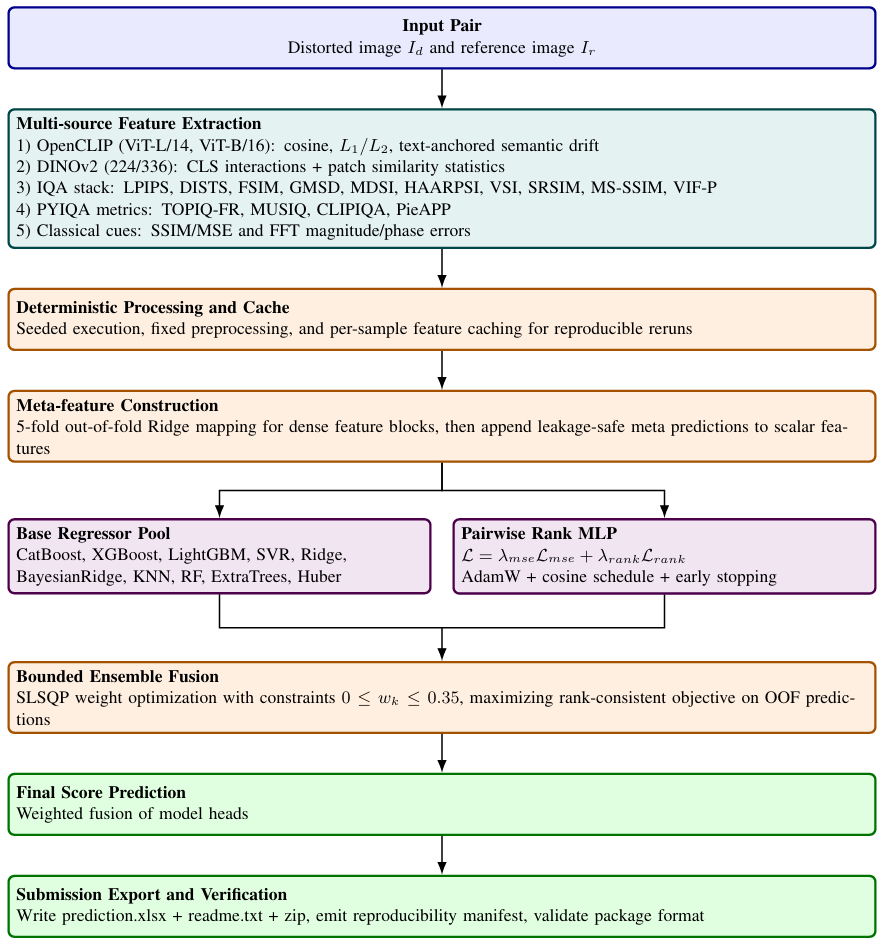}
    \caption{System diagram of Team Ayush Gupta}
    \label{fig:Ayush Gupta}
\end{figure}

This team proposes a pipeline following a comprehensive six-stage feature-based approach. First, the system extracts a diverse set of dense and scalar features from distorted and reference image pairs, incorporating semantic embeddings from OpenCLIP and DINOv2~\cite{radford2021learningclip,oquab2023dinov2}, perceptual metrics, and classical frequency-domain signals. Next, to effectively manage high-dimensional data, it constructs leakage-safe, out-of-fold meta-features from these dense blocks using Ridge regression. The core modeling phase then trains a heterogeneous pool of tabular base regressors (e.g., CatBoost, XGBoost, and LightGBM~\cite{prokhorenkova2018catboost,chen2016xgboost,ke2017lightgbm}) alongside a custom pairwise rank multilayer perceptron (MLP) designed to optimize relative ordering behavior. Finally, the framework employs a bounded weight optimization strategy to blend these diverse model heads based on out-of-fold predictions, ensuring rank-consistent fusion before retraining on the full dataset and exporting a strictly verified submission package.

The final system follows six stages:
\begin{enumerate}[1.]
\item Load ground-truth training labels.
\item Extract dense and scalar features for each pair.
\item Build out-of-fold meta-features from dense blocks.
\item Train diverse base regressors and a pairwise rank neural model.
\item Optimize bounded ensemble weights using OOF predictions.
\item Retrain on full data, infer on test pairs, and export submission package.
\end{enumerate}

\noindent\textbf{Training Details.} 
\begin{itemize}
\item Training: provided distorted-reference pairs with official scores.
\item Validation: internal 5-fold CV for model selection and blending.
\item Testing: hidden-score test pairs; final predictions exported in official format.
\end{itemize}
\begin{itemize}
\item Images loaded in RGB and downscaled to a max dimension cap for efficiency.
\item Standardized tensor conversion for IQA metrics.
\item Multiple resolutions used for feature diversity.
\item NaN-safe conversion and clipping where required.
\end{itemize}

\noindent\textbf{Testing Details.} 
At test time:
\begin{enumerate}[1.]
\item Extract/load cached features.
\item Generate meta predictions using full-data Ridge heads.
\item Scale features with full-data scaler.
\item Predict from all base models + rank model.
\item Fuse with optimized bounded weights.
\item Clip score range for safety and write official xlsx file.
\item Package with exact required readme format and zip structure.
\end{enumerate}

\noindent\textbf{Implementation Details.} 
The total method complexity is dominated by the parallel execution of large vision backbones (CLIP ViT-L/B, DINOv2-L) and IQA metric models. 
\begin{itemize}
\item \textbf{Total Parameters:} Approximately $\sim 1.2$ Billion parameters combined across all frozen feature extractors.
\item \textbf{GPU Memory Consumption:} Requires roughly $10-12$\,GB of VRAM during the batched feature extraction phase depending on resolution.
\item \textbf{Runtime:} Complete feature extraction and ensemble inference takes $\sim 1.5$ seconds per image pair on a modern consumer GPU (e.g., NVIDIA RTX 3090/4090).
\item \textbf{FLOPs:} High, dominated by the multi-scale ViT-L forward passes ($\sim$ hundreds of GFLOPs per pair).
\end{itemize}

\subsection{QA\_FTE}

\begin{figure}[t]
    \centering
    \includegraphics[width=1.0\linewidth]{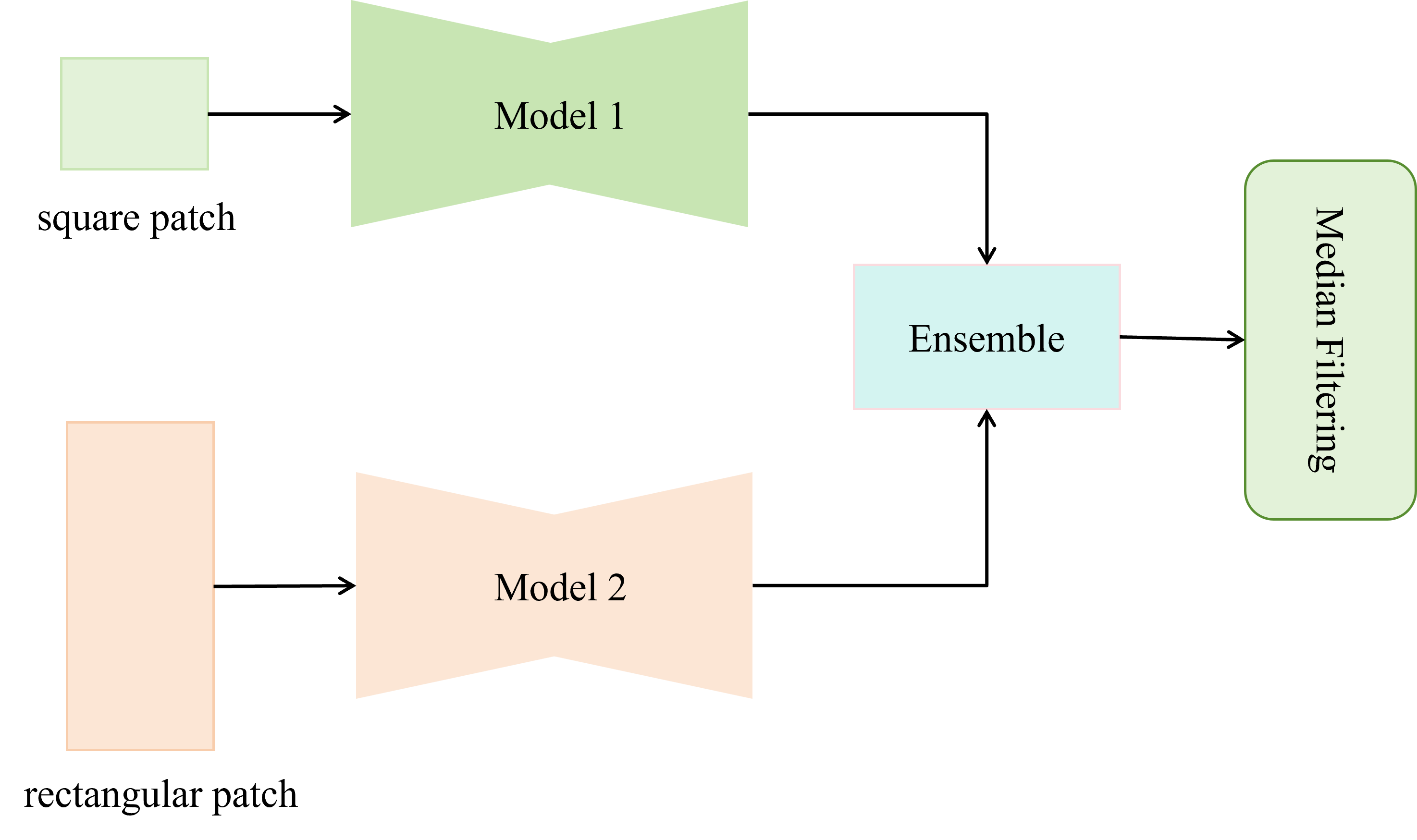}
    \caption{The pipeline of the method proposed by Team BUU\_CV}
    \label{fig:BUU_CV}
\end{figure}

This team proposes a reinforcement learning-based approach using Large Vision-Language Models (LVLMs) for Semantic Image Quality Assessment (SIQA). Specifically, they fine-tune the Qwen3-VL-8B-Instruct model~\cite{bai2025qwen3} using the Group Relative Policy Optimization (GRPO) framework. To elicit deeper reasoning and better alignment with human subjective judgments, they utilize a Chain-of-Thought (CoT) prompting strategy~\cite{wei2022chain} and Test-Time Scaling(TTS) method. The model is explicitly instructed to perform an internal reasoning process within \texttt{<thinking>} tags before predicting the final integer score.

\noindent\textbf{Training Details.} 
The model is trained using the VERL framework for GRPO optimization with a learning rate of $10^{-5}$ and a cosine warmup schedule. A Strict Binary Reward mechanism is implemented, where the model receives a reward only if the predicted integer score exactly matches the ground truth. And the KL divergence constraint is ignored due to the limited size of training data.The training is conducted for 50 episodes with a mini-batch size of 8 per GPU, sampling $N=32$ responses per prompt to compute group relative advantages.

\noindent\textbf{Testing Details.} 
During inference, a Test-Time Scaling (TTS) strategy is adopted. For each test image pair, the model samples $K=64$ responses at a temperature of 1.0. The final semantic quality score is obtained by averaging the valid extracted scores from these multiple reasoning pathways, which helps to improve robustness and reduce variance.

\noindent\textbf{Implementation Details.} 
The method utilizes the Qwen3-VL-8B-Instruct (approx. 8B parameters) as the backbone architecture. The maximum prompt length is set to 4096 tokens and the response length to 2048 tokens. The implementation uses vLLM for accelerated generation with prefix caching and chunked prefill enabled. No extra training data beyond the provided LoViF dataset is used.

\subsection{RuntimeTerror}

This team proposes a hybrid semantic quality assessment framework that combines high-level semantic reasoning features with low-level deterministic perceptual features, followed by an ensemble-based regression model. Specifically, they extract structured semantic features using a 4-bit quantized Qwen-based vision-language model~\cite{bai2025qwen3} to decompose semantic similarity into interpretable components like subject consistency, spatial layout and object relationships. Simultaneously, deterministic metrics such as LPIPS~\cite{zhang2018lpips} and SSIM~\cite{wang2004imagessimloss} are computed using CLIP embeddings~\cite{radford2021learningclip} to capture low-level artifacts and visual fidelity. To capture complex cross-domain interactions, they perform explicit feature mixing using arithmetic operations (addition, subtraction, multiplication), resulting in a compact 18-dimensional feature vector.
\begin{equation}
    x_{ij}^{(add)}=x_{i}+x_{j},\quad x_{ij}^{(sub)}=x_{i}-x_{j},\quad x_{ij}^{(mul)}=x_{i}\cdot x_{j}
\end{equation}
 Finally, this representation is fed into an ensemble of CatBoost regressors~\cite{prokhorenkova2018catboost} to predict the continuous quality score.

\noindent\textbf{Training Details.} 
The models are trained exclusively on the provided LoViF dataset without any external labeled data. The 18-dimensional features are precomputed and normalized where required. The regression task is optimized to predict continuous quality scores using an ensemble of CatBoost models, where multiple independent regressors are trained with different seeds to capture diverse feature interactions and decision boundaries.The final prediction is computed as:
\begin{equation}
    \hat{y}_{\text{ensemble}} = \frac{1}{K} \sum_{k=1}^{K} f^{(k)}(x)
\end{equation}

\noindent\textbf{Testing Details.} 
During inference, the same feature extraction and mixing pipeline is applied to the test image pairs. Each independent CatBoost model in the ensemble generates a score prediction. The final semantic quality score is obtained through ensemble averaging, which significantly reduces prediction variance and enhances robustness to subjective quality labels.

\noindent\textbf{Implementation Details.} 
The team relies on a 4-bit quantized Qwen model for semantic features and a CLIP model for deterministic embeddings. They jointly leverage
 semantic features based on vision-language and deterministic perceptual metrics, capturing differences in meaning-level and visual fidelity.The final dimension of the fused feature is strictly limited to 18. CatBoost is chosen as the regression backbone due to its effectiveness on low-dimensional structured tabular data and its ability to handle heterogeneous features without strict normalization.

\subsection{DSS-SQA}
\begin{figure}[t]
    \centering
    \includegraphics[width=1.0\linewidth]{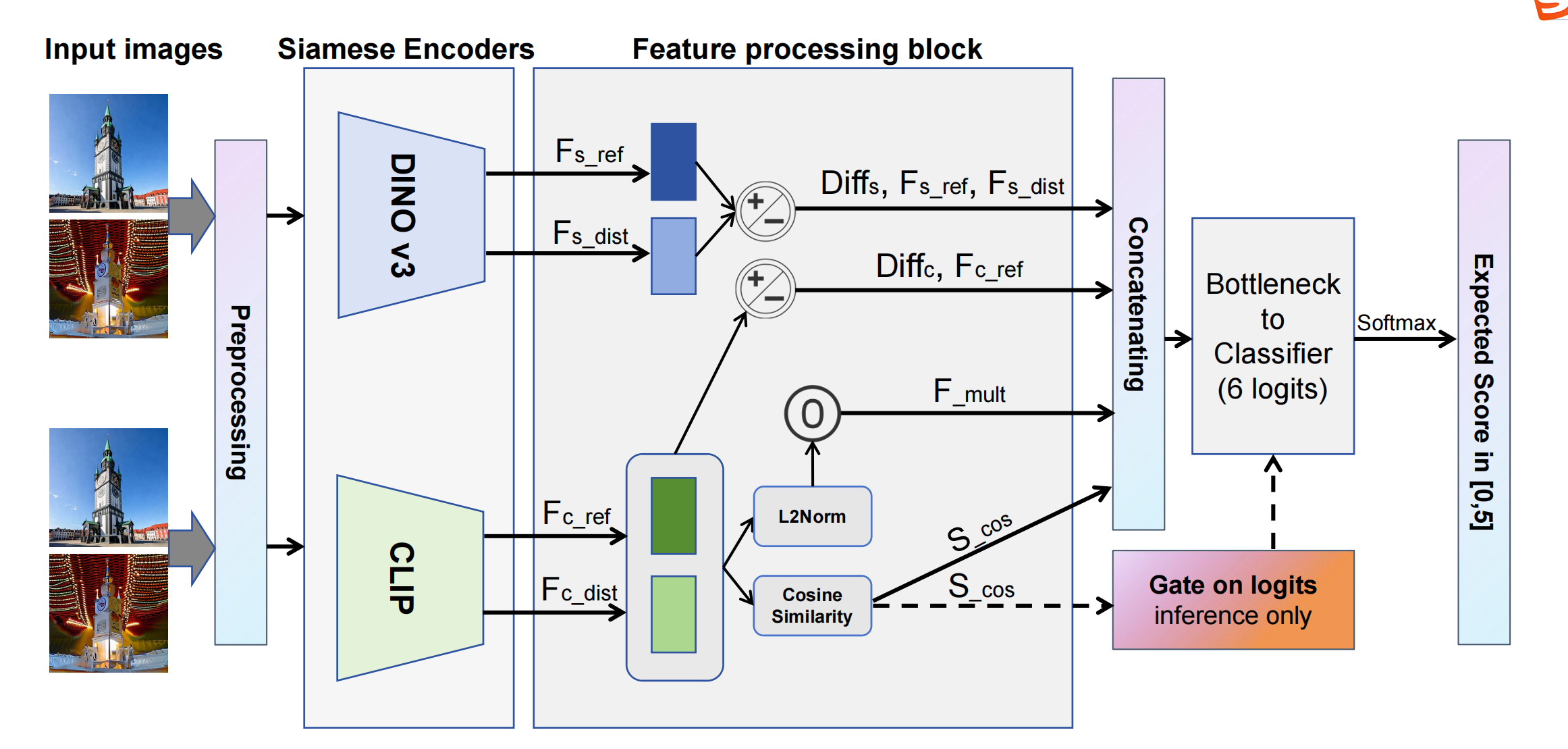}
    \caption{ The main structure of Team DSS-SQA}
    \label{fig:DSS-IQA}
\end{figure}
This team proposes a Siamese architecture~\cite{bromley1993siamese} (as the Figure~\ref{fig:DSS-IQA} shows), named DSS-SQA, that explicitly decouples structural fidelity from semantic alignment to assess generative visual quality. The model employs a dual-vision backbone featuring a frozen DINOv3-Base transformer from the DINO family~\cite{oquab2023dinov2} for structure-aware global representations and a CLIP-ViT-L/14 model~\cite{radford2021learningclip} for human-aligned semantic embeddings. To mitigate shortcut learning on limited data, they introduce an Element-wise Multiplication Fusion for CLIP features. Specifically, features from reference and distorted images are $L_2$ normalized before computing their Hadamard product.
\begin{equation}
mult_clip = L_2(F_{c,ref}) \odot L_2(F_{c,dist})
\end{equation}
The final fused representation—comprising original features, absolute differences, the multiplication result, and scalar cosine similarity ($S_{cos}$)—is processed through a bottleneck layer and a classification head to output discrete quality category logits.

\noindent\textbf{Training Details.}
The backbone models are kept frozen, and only the bottleneck and prediction head (~1.41M parameters) are trained using the AdamW optimizer with a cosine annealing schedule. The training strictly uses the 510 provided image pairs and employs a joint loss function combining Cross-Entropy (CE) for classification and Mean Squared Error (MSE) for score regression.

\noindent\textbf{Testing Details.}
During inference, the model derives a continuous quality score as the mathematical expectation of the Softmax distribution over the 6 output logits. Crucially, an Explicit Semantic Gating mechanism is applied: if $S_{cos}$ falls below a threshold of 0.4, the gate forcefully pushes the logits toward the lowest quality class (Class 0) to penalize semantic hallucinations.

\noindent\textbf{Implementation Details.}
The framework is implemented using DINOv3-Base (~85.6M) and CLIP-ViT-L/14 (~303.5M). Images are resized to $224\times224$ and internally interpolated to $336\times336$ for the CLIP branch. To ensure robustness, a 5-fold stratified cross-validation strategy is utilized, where the final prediction is an ensemble average of 5 independent models

\subsection{cythdg}
\begin{figure}[t]
    \centering
    \includegraphics[width=0.5\linewidth]{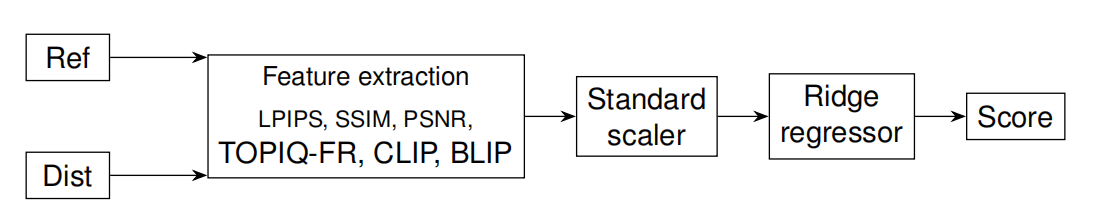}
    \caption{Pipeline of the method proposed by Team cythdg}
    \label{fig:cythdg}
\end{figure}
This team proposes a lightweight full-reference (FR) pipeline (as Figure ~\ref{fig:cythdg} illustrates)  that maps a compact combination of hand-crafted perceptual metrics and pretrained semantic features to a quality score using Ridge regression. The model extracts low-level and perceptual features using LPIPS~\cite{zhang2018lpips}, SSIM~\cite{wang2004imagessimloss}, PSNR, and the learned TOPIQ-FR metric from PyIQA~\cite{gu2023pyiqa}. To account for high-level semantic consistency, they optionally incorporate reference-distortion cosine similarity from frozen vision-language encoders, specifically CLIP ViT-B-32~\cite{radford2021learningclip} and BLIP~\cite{li2022blip}. All extracted features are standardized to zero mean and unit variance before being mapped to a scalar quality score in the range [0, 5].
\begin{equation}
Score = \mathbf{w}^T \mathbf{x} + b, \quad \text{s.t.} \quad \min_{\mathbf{w}, b} | \mathbf{y} - (\mathbf{Xw} + b) |^2_2 + \alpha | \mathbf{w} |^2_2
\end{equation}
The final prediction is generated by a single linear model, ensuring high interpretability and minimal computational overhead.

\noindent\textbf{Training Details.}
The model is trained exclusively on the provided competition training set. The training process is deterministic and closed-form, utilizing 5-fold cross-validation over a small grid of candidates to determine the optimal regularization strength ($\alpha$). No end-to-end deep network is trained; only the Ridge coefficients and intercept are optimized while all feature extraction backbones remain frozen.

\noindent\textbf{Testing Details.}
During inference, reference and distorted images are processed through the same one-stage feature extraction pipeline. The features are standardized using the parameters learned during training and then fed into the fitted Ridge regressor to produce a continuous quality score. The solution avoids any ensemble or fusion strategies to maintain simplicity and full reproducibility.

\noindent\textbf{Implementation Details.}
The implementation relies on the \textit{pyiqa} toolkit~\cite{gu2023pyiqa} for standard IQA metrics and TOPIQ-FR, alongside OpenCLIP and Hugging Face transformers for semantic similarity. The resulting model size is extremely compact, on the order of a few kilobytes, making it highly suitable for efficient deployment. The computational cost scales linearly with the number of images, as it is dominated by the off-the-shelf feature extractors.

%% file: sec/5_Appendix.tex
\appendix
\section*{Organizers and Teams}
\label{organizers}

\subsection*{Organizers}

\noindent\textit{\textbf{Title:}} LoViF 2026 Challenge on Human-oriented Semantic Image Quality Assessment @ CVPR2026

\noindent\textit{\textbf{Members:}} 

Xin Li\textsuperscript{1} (\textcolor{magenta}{xin.li@ustc.edu.cn})

Daoli Xu\textsuperscript{2} (\textcolor{magenta}{xudaoli@emergetech.com.cn})

Guoqiang Xiang\textsuperscript{2} (\textcolor{magenta}{
xiangguoqiang@emergetech.com.cn})

Zhibo Chen\textsuperscript{1} (\textcolor{magenta}{chenzhibo@ustc.edu.cn})

Jian Guan \textsuperscript{2} (\textcolor{magenta}{guanjian@emergetech.com.cn})

Weiping Li  \textsuperscript{1} (\textcolor{magenta}{wpli@ustc.edu.cn})

\noindent\textit{\textbf{Affiliations:}}

\noindent\textsuperscript{1} University of Science and Technology of China

\noindent\textsuperscript{2} EMERGETECH

\subsection*{RedpanQA Alliance}

\noindent\textit{\textbf{Title:}} Reference-Based Semantic Quality Assessment with Multimodal Large
Language Models

\noindent\textit{\textbf{Members:}} Weixia Zhang\textsuperscript{1}(\textcolor{magenta}{lirunzhe@stu.hit.edu.cn}), Wei Sun\textsuperscript{2}, Zhihua Wang\textsuperscript{3}, Dandan Zhu\textsuperscript{2}, Chengguang Zhu\textsuperscript{4}

\noindent\textit{\textbf{Affiliations:}}

\noindent  \textsuperscript{1} School of Computer Science, Shanghai Jiao Tong University

\noindent  \textsuperscript{2} School of Computer Science and Technology, East China Normal University

\noindent  \textsuperscript{3} School of Cyber Science and Technology, Sun Yat-sen University

\noindent  \textsuperscript{4} Changzhou Microintelligence Co.,Ltd

\subsection*{Ayush Gupta}

\noindent\textit{\textbf{Title:}} Rank-Consistent Multi-Feature Ensemble for Semantic Image Quality
Assessment

\noindent\textit{\textbf{Members:}} Ayush Gupta
(\textcolor{magenta}{ayushgupta123111@gmail.com})

\noindent\textit{\textbf{Affiliations:}}
\noindent Netaji Subhas University of Technology

\subsection*{RuntimeTerror}

\noindent\textit{\textbf{Title:}} Joint Modeling of Semantic Consistency and Visual Fidelity for Image Quality
Assessment

\noindent\textit{\textbf{Members:}} Rachit Agarwal (\textcolor{magenta}{rachit.a@samsung.com}), Shouvik Das

\noindent\textit{\textbf{Affiliations:}} Samsung R\&D Institute, Bengaluru, India

\subsection*{QA-FTE}

\noindent\textit{\textbf{Title:}} QA-FTE: Chain-of-Thought Reasoning and Test-Time Scaling in Large Vision-Language Models for Semantic Image Quality Assessment

\noindent\textit{\textbf{Members:}} Kanglong Fan\textsuperscript{1} (\textcolor{magenta}{kanglofan2-c@my.cityu.edu.hk}), Wen Wen\textsuperscript{2}, Shuyan Zhai\textsuperscript{1}, Tianwu Zhi\textsuperscript{2}, Aoxiang Zhang\textsuperscript{2}, Jianzhao Liu\textsuperscript{2}, Yabin Zhang\textsuperscript{2}

\noindent\textit{\textbf{Affiliations:}}

\noindent  \textsuperscript{1} City University of Hong Kong

\noindent  \textsuperscript{2} ByteDance Inc.

\subsection*{DSS-SQA}

\noindent\textit{\textbf{Title:}} DSS-SQA: Decoupling Structure and Semantics for Semantic Quality Assessment

\noindent\textit{\textbf{Members:}} Jiajun Wang\textsuperscript{1} (\textcolor{magenta}{jiajun.wang@fau.de}), Yipeng Sun\textsuperscript{1}, Kaiwei Lian\textsuperscript{2}

\noindent\textit{\textbf{Affiliations:}} 

\noindent \textsuperscript{1} Pattern Recognition Lab, Friedrich-Alexander Universität Erlangen-Nürnberg Martensstraße 3, 91058 Erlangen, Germany

\noindent \textsuperscript{2} University of Manchester,
Oxford Road, Manchester, M13 9PL, United Kingdom

\subsection*{cythdg}

\noindent\textit{\textbf{Title:}} FR-IQA Features and Ridge Regression for Semantic Quality Prediction

\noindent\textit{\textbf{Members:}} Banghao Yin (\textcolor{magenta}{ybh666@mail.ustc.edu.cn})

\noindent\textit{\textbf{Affiliations:}} School of Computer Science and Technology, University of Science and Technology of China, Hefei, China